# Improved dual channel pulse coupled neural network and its application to multi-focus image fusion

Huai-Shui Tong [1, 2], Xiao-Jun Wu [1, *], Hui Li [1]

1 School of IoT Engineering, Jiangnan University, Wuxi 214122, P R China；

2 College of Mathematics and Information Science, East China Institute of Technology, Fuzhou, 344000, P R China

Abstract：This paper presents an improved dual channel pulse coupled neural network (IDC-PCNN) model for image fusion. The model can overcome some defects of standard PCNN model. In this fusion scheme, the multiplication rule is replaced by addition rule in the information fusion pool of dual channel PCNN (DC-PCNN) model. Meanwhile the sum of modified Laplacian (SML) measure is adopted, which is better than other focus measures. This method not only inherits the good characteristics of the standard PCNN model but also enhances the computing efficiency and fusion quality. The performance of the proposed method is evaluated by using four criteria including average cross entropy, root mean square error, peak value signal to noise ratio and structure similarity index. Comparative studies show that the proposed fusion algorithm outperforms the standard PCNN method and the DC-PCNN method.

Keywords：IDC-PCNN, Image fusion, PCNN, DC-PCNN, SML

## 1. Introduction

PCNN model was firstly proposed by Eckhorn et al in 1990 [1]. The inherent ability to feature extraction, denoising, and segmentation among others is a very important property of the PCNN model, which makes the model a very powerful image-processing tool. In recent years, PCNN model and its improved models have been widely applied to different tasks of image processing, for instance, image denoising, image segmentation, and some other fields, especially in image fusion. However, it is difficult to make mathematical analysis of PCNN model because it is a nonlinear one. Furthermore, there are many uncertain parameters in PCNN model. Therefore, an IDC-PCNN model is proposed in this paper to overcome the aforementioned defects. In this model, the internal state of the neuron is combined by addition rule rather than multiplication rule of the two input neurons, and the SML is adopted to construct weighted coefficients of the model. Furthermore, the computational efficiency is improved in the proposed model.

As an application to the IDC-PCNN model, the multi-focus image fusion is introduced. The optical lens of camera is limited by the focal length, thus camera cannot take pictures everywhere in focus. Some objects cannot be in focus, which leads to blurred image blocks. Image fusion technique is usually used to solve the problem. Some computationally simple methods are proposed in [2][19][20][21][22][25][28][29], and the novel feature extraction methods are also introduced [23][24][26][27][30]. In literature [3], the artificial neural network based approach is presented. Authors claimed that their method is better than the wavelet transform based method,

---

*Corresponding author.

E-mail address: wu_xiaojun@jiangnan.edu.cn, xiaojun_wu_jnu@163.com (Xiao-Jun Wu).

especially when there exist movement object or misregistration. The standard coefficient combining methods is to compare focus measure of the source images and find the best focus one at the particular pixel location. The support vector machine approach outperforms these conventional schemes both quantitatively and visually [4]. The redundant wavelet transform method was presented in literature [5], which is a computational efficient method. Furthermore, the blurring effects, misregistration and sensitivity to noises have been effectively overcome by the method partially. In literature [6], the estimation of the noise strength and the pre-fixation of the window size are effectively avoided via the MW technique. The differential evolution algorithm is reliable, stable, robust and sensitive to the initial value and the control parameters [7]. The Fuzzy C-Mean (FCM) algorithm is proposed to partition the feature region of the image in literature [8], and to calculate the distance of the regions as regional dissimilarity. This regional dissimilarity is benefit for human visual perception and can obtain good fusion effect by using to multi-focus image fusion.

Recently, the fusion methods based on PCNN have been developed. A modified approach of PCNN suitable for application in image fusion technique is proposed in literature [9], which can improve the computational efficiency and reduce the processing time. In literature [10], the linking strength is the clarity of each pixel, and the time matrix of the sub-images is obtained by the synchronous pulse burst property of the neurons. Fusing the time matrix and linking strength obtain the better fusion effect. In literature [11], the input source images are decomposed into several blocks to compute energy of Laplacian as feature maps. The outputs can be obtained by inputting the feature maps as the external stimulus. The fusion image is constructed by selecting the source image blocks that are obtain by comparing the outputs of images. Each source image must be inputted to the PCNN model, which brings large amount of calculation and complex model parameters. While the selection of the model parameters is a significant problem of PCNN model. Thus, model modification and parameters optimization are worthy to be studied. Therefore, an improved method is proposed in this paper. The method replaces multiplication rule with addition rule in the signal modulation process, SML measure is adopted to construct weighted coefficient and the active factor is removed. Performances show that our method enhances the computing efficiency, and gets better fusion quality. The improved DC-PCNN model based fusion algorithm is drawn in Fig.1. The main components of Fig.1 will be described in Section 3.

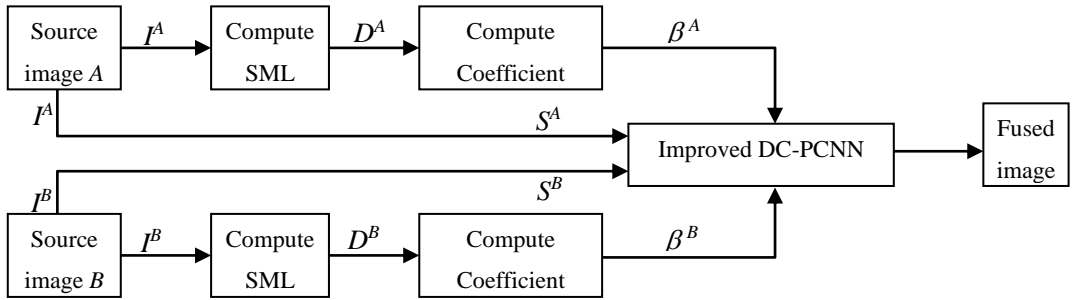

Fig.1 The improved DC-PCNN based fusion algorithm

The remainder of this paper is presented as follows. The standard PCNN and DC-PCNN is reviewed in section 2.1, and the IDC-PCNN is presented in detail in section 2.2. The proposed fusion algorithm of this paper is given in Section 3. The experiment and performance evaluation are given in Section 4 and the conclusions are given in Section 5.

## 2. Pulse coupled neural network model

Since PCNN and DC-PCNN are foundations of the proposed model in this paper, they are reviewed in section 2.1. Then the proposed IDC-PCNN is presented in section 2.2.

**2.1 PCNN and DC-PCNN model**

Neurons of standard PCNN model are consisting by three components. One is the dendritic tree, which is to receive the signals from the feeding field and the linking field. The other is linking modulation, which is the unit to combine the signals from the feeding and linking. The pulse generator is the part to generate pulse. The neurons receive the input signals such as pulses, constants, or external stimuli. This input signals come from the neighbor neurons or the other inputs. While the adaptation processes of PCNN model is fully compatible and complementary. The network's nonadaptive spatiotemporal dynamics of the PCNN is the main focus of interest [11]. More description of PCNN model can be found in the literatures [11,12].

Although PCNN model is useful in image processing, there are some defects exist in the standard PCNN model. It is difficult to make mathematical analysis of PCNN model because it is a nonlinear one. Furthermore, there are many uncertain parameters in PCNN model. Some work is done to overcome these shortcomings. Wang and Ma [13,14] proposed a DC-PCNN model to overcome some defects of PCNN. However, the DC-PCNN model inherits the nonlinear property of PCNN, because the two input neurons must be multiplied. As example, the DC-PCNN model was applied to multi-focus image fusion. The two source images are taken as input neurons, and the neurons are combined with multiplication rule to internal state in the fusion pool. By inputting a DC-PCNN model, which has fewer parameters and higher computing efficiency and better fusion quality compared to PCNN model [13], the two source images are fused.

The DC-PCNN model (see Fig.2) consists of three parts. The first part is the dendritic tree. The external stimuli and surrounding stimuli will be inputted into the first part. The information fusion pool is the second part. All signals will be coupled in this part. The last part is the pulse generator. The output pulse will be generated in the part.

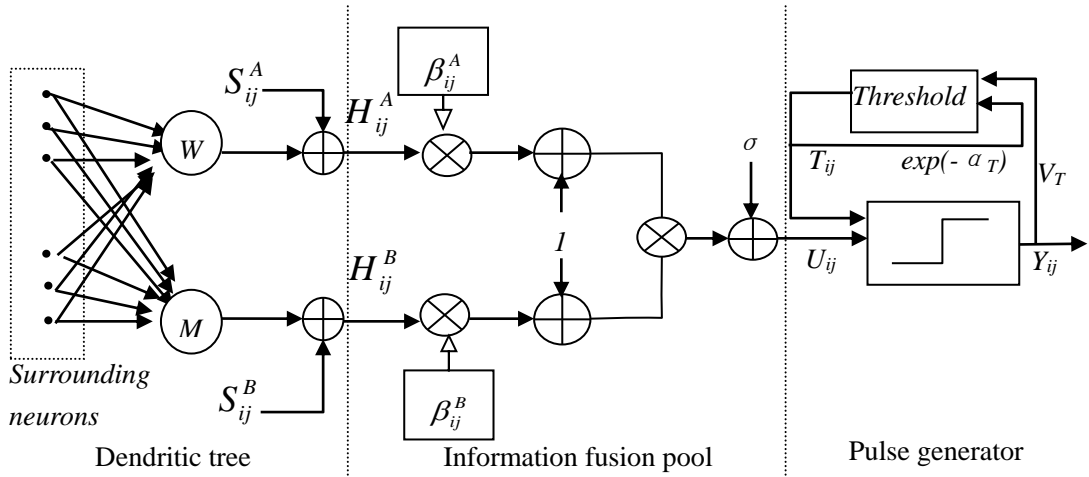

Fig.2 The structure of DC - PCNN

The equations of the DC-PCNN model and more details about the algorithm can be found in the literature [13].

**2.2 IDC – PCNN model**

From Fig.2 we can see that the combined internal state is generated by multiplying the information of two channels, and by adding a level factor. Apparently, there are two shortcomings of DC-PCNN. The first one is that the computational load is high. The second one is that it is difficult to determine the level factor. Motivated by these observations, we modify the structure of DC-PCNN by replacing multiplication with addition, and omit the parameter of level factor.

Again, the IDC-PCNN model (see Fig.3) is also composed of the following components: dendritic tree, image fusion pool and pulse generator, respectively.

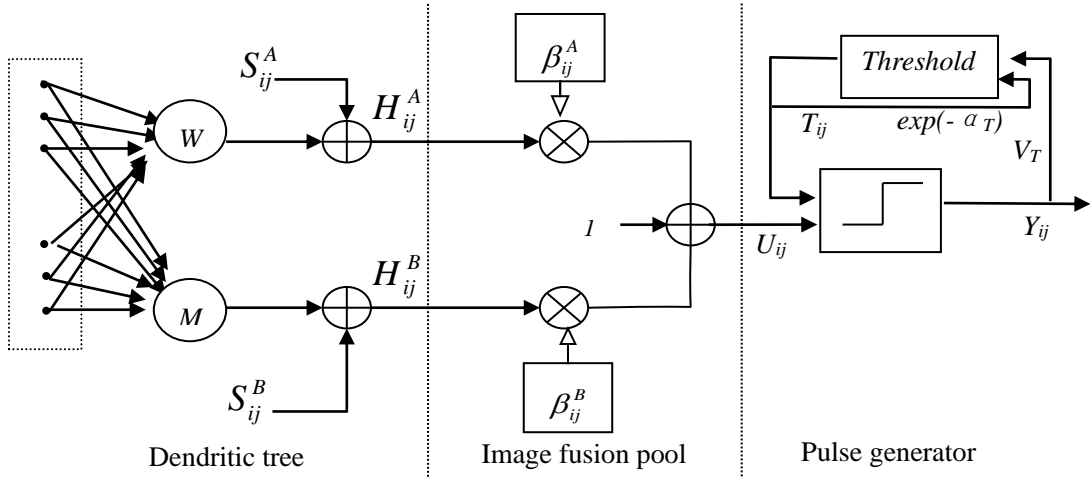

Fig.3 The structure of the IDC-PCNN

In the DC-PCNN model, as shown in Fig.2, each iteration process requires three times calculation of addition and multiplication. Suppose the images size is $m \times n$, and $N$ is the number of iteration, then the total computation of information fusion in the DC - PCNN model is $3 \times N \times m \times n$ times of multiplication and addition. However, for the improved model, as shown in Fig.3, each iteration process needs only one addition and twice multiplications calculation in image fusion pool, and thus the computation is $2 \times N \times m \times n$ times multiplication and $N \times m \times n$ times addition. Compared with DC - PCNN, our model improves the computational efficiency.

Some good features of the PCNN model are inherited in the IDC-PCNN model. Input stimulus and stimulus from neighbors are mathematically shown in Eq. (1) – (2). The neuron fires when the combined internal state value rises greater than the threshold value. Then there is an iteration process. The threshold value is increased significantly and it decays until the internal activity value rises greater than the threshold value again. This iteration process leads to the dynamic pulse burst nature of the model. Similarly, the other properties of PCNN are also remained in the IDC-PCNN model.

There are two input channels in the improved model. The stimulus can be inputted into the two channels at the same time. The mathematical model is given as the following equations:

$$\begin{cases} H_{ij}^{A}(n) = S_{ij}^{A} + \sum_{k,l} w_{ijkl} Y_{kl}(n-1) & (1) \\ H_{ij}^{B}(n) = S_{ij}^{B} + \sum_{k,l} m_{ijkl} Y_{kl}(n-1) & (2) \\ U_{ij}(n) = 1 + \beta_{ij}^{A} H_{ij}^{A}(n) + \beta_{ij}^{B} H_{ij}^{B}(n) & (3) \\ Y_{ij}(n) = \begin{cases} U_{ij}(n) - R_{ij}(n) - 1, & U_{ij}(n) > T_{ij}(n-1) \\ 0, & \text{otherwise} \end{cases} & (4) \\ T_{ij}(n) = \begin{cases} e^{-\alpha_T} T_{ij}(n-1), & Y_{ij}(n) = 0 \\ V_T, & \text{otherwise} \end{cases} & (5) \end{cases}$$

The two symmetrical input channels are given by Eq. (1)-(2), where $S^A$ and $S^B$ are external input stimulus. $w_{ijkl}$ and $m_{ijkl}$ are the two synaptic weighting coefficient of the neuron at $(i,j)$. The image fusion pool is given by Eq. (3), and $U_{ij}$ is the coupled internal activity of the neuron. Where $\beta^A$ and $\beta^B$ are the weighting coefficients. The pulse generator is given by Eq. (4)-(5). $T_{ij}$ is the dynamic threshold. $Y_{ij}$ is determined by $U_{ij}$ and $T_{ij}$. $\alpha_T$ is the time constant and $V_T$ is the normalized constant. $R$ denotes the combination of the surrounding neurons. $\mu$ is the connection coefficient. Usually,

$$\mu_{ijkl} = w_{ijkl} = m_{ijkl}, \qquad R_{ij} = \sum_{k,l} \mu_{ijkl} Y_{kl}(n-1)$$

This model is almost the same as the DC-PCNN model, while the only difference is given by Eq. (3). The multiplication rule is replaced by addition rule, and active factor σ is removed in our model. The computational efficiency is improved by this replacement and the model is simplified by removing the active factor, which can be seen in section 4. In addition, the weighting coefficients are determined via experiment analyses, which can be seen in section 3.2.

The algorithm of IDC-PCNN model is summarized as follows:
1) Initialization.
[m n] = size(image), U = O = Y =zeros(m,n), T =ones(m,n), K = [1,0.5, 1; 0.5, 0,0.5; 1,0.5, 1];
Where O is the output. K is the connection coefficient.
2) Normalize the external stimuli.
$S^A$ =Normalized ($S^A$), $S^B$ =Normalized ($S^B$);
3) do
    R=Y⊙K;
    $H^A = S^A + R$;
    $H^B = S^B + R$;
    For ( i=1,…m,j=1, …n)
       $U_{ij} = 1 + \beta^A{}_{ij} \times H^A{}_{ij} + \beta^B{}_{ij} \times H^B{}_{ij}$;
       If $U_{ij} > T_{ij}$    then    $Y_{ij} = U_{ij} - R_{ij} - 1$, else $Y_{ij} = 0$;
       If $S^A{}_{ij} = S^B{}_{ij}$ or $\beta^A{}_{ij} = \beta^B{}_{ij}$, then $O_{ij} = S^A{}_{ij}$ or $S^B{}_{ij}$; else $O_{ij} = Y_{ij}$;
       If $Y_{ij} = 0$ then $T_{ij} = exp(-\alpha_T) \times T_{ij}$, else $T_{ij} = V_T$.
    End for
  Until (all neurons have been fired).

4) Output $O$.

## 3. Image fusion algorithm

A fusion algorithm based on the improved model is given in first subsection. Moreover, the weighting coefficient is an important parameter of the improved model. Therefore, some evaluation criteria for image sharpness are discussed, and a method to obtain the weighting coefficient by SML is described in the next subsection.

### 3.1 Fusion algorithm

PCNN based image fusion method requires no training, and the source images should be registered. Two source images are denoted as $S^A$ and $S^B$, then input the external stimuli $S^A(i,j)$ and $S^B(i,j)$ to the IDC-PCNN model, then the fusion image is obtained when all neurons have been fired. Fig.1 shows the main components of our algorithm.

The proposed fusion method is described as four steps:

(1) Compute SML with Eq. (6) for each source image. Denote $D^A$ and $D^B$ as the measured images, respectively.

(2) Calculate the weighting coefficients $\beta^A$ and $\beta^B$ via $D^A$ and $D^B$ using Eq.(8) and Eq.(9).

(3) Input the two stimuli $S^A$ and $S^B$ into the improved model, and then start the model.

(4) Obtain the fused image via the IDC-PCNN.

### 3.2 Weighting coefficient

Recently, many sharpness assessment criteria are discussed. Typical focus measure methods are: variance (Var), spatial frequency (SF), energy of Laplacian (EOL), and sum of modified Laplacian (SML), etc. These evaluation indices can be used to describe the image sharpness individually, while SML and EOL can usually present better effect than the others [15]. EOL is the sharpness assessment criterion to measure image focus and construct the weighting coefficient. However, experiments reveal that different measure used in DC-PCNN model will produce different fusion result, which can be seen in Fig.4. We use SML measure to construct the weight coefficient, which is the best fusion performance of the four mentioned measures. Experiment results in table 2 and table 4 verify the statement, but there is no significant difference. Thus, the similar results are obtained by using the four measures in our proposed model, but the different results are obtained by using the four measures in DC-PCNN model, especially the RMSE value in Table 3. The mathematic expression of SML is given as follows:

$$L(i,j) = \sum_{x=i-N}^{i+N} \sum_{y=j-N}^{j+N} \nabla_{ML}^2 I(x,y), \qquad \nabla_{ML}^2 I(x,y) \geq T$$

$$\nabla_{ML}^2 I(x,y) = \left| -I(x-k,y) + 2I(x,y) - I(x+k,y) \right| \qquad (6)$$
$$+ \left| -I(x,y-k) + 2I(x,y) - I(x,y+k) \right|$$

Where $k$ is usually equal to $1$, $T$ is a threshold, and $N$ determines the window size to compute the focus measure. $L(i,j)$ is the element of the SML, $I(x,y)$ is the image pixel value. The construction of weighting coefficient is described in detail in literature [13]. We introduce it

briefly as follows:

1) Suppose source images $A(i,j)$ and $B(i,j)$, transforming into measured images $D^A(i,j)$ and $D^B(i,j)$ after focus measure SML is used in this paper;

2) Define $M(i,j)=D^A(i,j)–D^B(i,j)$.

3) Summing all the point around the decision point in the $(r+1) \times (r+1)$ region, we obtain

$$\overline{M}(i,j) = \sum_{x=-r/2}^{r/2} \sum_{y=-r/2}^{r/2} M(i+x, j+y) \qquad (7)$$

4) The weighting coefficients are determined by

$$\beta_{ij}^A = \frac{1}{1+e^{-\eta \overline{M}(i,j)}} \qquad (8)$$

and

$$\beta_{ij}^B = \frac{1}{1+e^{\eta \overline{M}(i,j)}} \qquad (9)$$

## 4. Experiments and Analysis

The experiments and analysis are done to confirm the validity of our algorithm. The computer is HP ProBook 4321s, and Matlab7.6.0 (R2008a), which is the experimental environment in Windows 7. The neighborhood window width parameter N is equal to 2, and the threshold parameter T is equal to zero, and the other parameters are same with the Wang's method [13]. The performance is done to contrast our method to Huang's method [11] and Wang's method [13]. The objective evaluation criteria are average cross entropy (CE), root mean square error (RMSE), peak signal to noise ratio (PSNR) and structural similarity index (SSIM) [16]. The mathematical expressions of the measures are given as follows:

$$CE_{AB} = \sum_{i=0}^{L-1} p_i \log_2 \frac{p_i}{q_i}, \qquad CE = (CE_{AF} + CE_{BF})/2 \qquad (10)$$

$$RMSE = \sqrt{\frac{\sum_{i=1}^{m} \sum_{j=1}^{n}[F(i,j) - R(i,j)]^2}{m \times n}} \qquad (11)$$

$$PSNR = 10\log_{10}(255 * 255 / RMSE^2) \qquad (12)$$

$$SSIM = \frac{(2\mu_R \mu_F - C_1)(2\sigma_{RF} + C_2)}{(\mu_R^2 + \mu_F^2 + C_1)(\sigma_R^2 + \sigma_F^2 + C_2)} \qquad (13)$$

Where the source images are $A$ and $B$, $F$ is the fused image and $R$ is the reference image. $p_i$ and $q_i$ are distribution probability of image pixels value i. The CE is image average cross entropy, and CE directly reflects the corresponding pixel difference of the source images. The smaller CE implies that the fusion method extracts the more information from the source images. The RMSE is a criterion to quantify differences between two images. Small RMSE means better effect for the fusion image. The SSIM is first given by Wang et al [16], and the program can be downloaded in [17]. The SSIM denotes the similarity of the fusion image and the reference image. Larger value shows better effect of the fusion image.

### 4.1 Experiments of weighting coefficient

According to the section 3.2, experiments were done. The representative results are given in Table 1 through Table 4 and Fig.4.

Table 1. Evaluation of different measures used in DC - PCNN (clock)

|     | CE     | RMSE    | PSNR    | SSIM   |
|-----|--------|---------|---------|--------|
| Var | 0.2717 | 12.3688 | 26.2842 | 0.8635 |
| SF  | 0.3145 | 12.9680 | 25.8733 | 0.8675 |
| SML | 0.2717 | 12.3688 | 26.2842 | 0.8635 |
| EOL | **0.2417** | **11.0804** | **27.2397** | **0.8753** |

Table 2. Evaluation of different measures used in improved DC - PCNN (clock)

|     | CE     | RMSE    | PSNR    | SSIM   |
|-----|--------|---------|---------|--------|
| Var | 0.2426 | 11.9811 | 26.5609 | 0.8202 |
| SF  | 0.2724 | 10.9675 | 27.3286 | 0.8835 |
| SML | **0.2366** | **10.8835** | **27.3955** | **0.8861** |
| EOL | 0.2416 | 11.0801 | 27.2399 | 0.8754 |

Table 3. Evaluation of different measures used in DC - PCNN (TSINGHUA)

|     | CE     | RMSE    | PSNR    | SSIM   |
|-----|--------|---------|---------|--------|
| Var | 1.7059 | 65.6741 | 11.7829 | 0.7139 |
| SF  | 1.6955 | 68.1093 | 11.4667 | 0.6973 |
| SML | 1.7059 | 65.6741 | 11.7829 | 0.7139 |
| EOL | **1.2824** | **8.6508** | **29.3897** | **0.9587** |

Table 4. Evaluation of different measures used in improved DC - PCNN (TSINGHUA)

|     | CE     | RMSE   | PSNR    | SSIM   |
|-----|--------|--------|---------|--------|
| Var | 0.0033 | 0.8382 | 49.6639 | 0.9987 |
| SF  | 0.0031 | 0.5531 | 53.2748 | 0.9994 |
| SML | **0.0031** | **0.5490** | **53.3393** | **0.9994** |
| EOL | 0.0031 | 0.5684 | 53.0383 | 0.9994 |

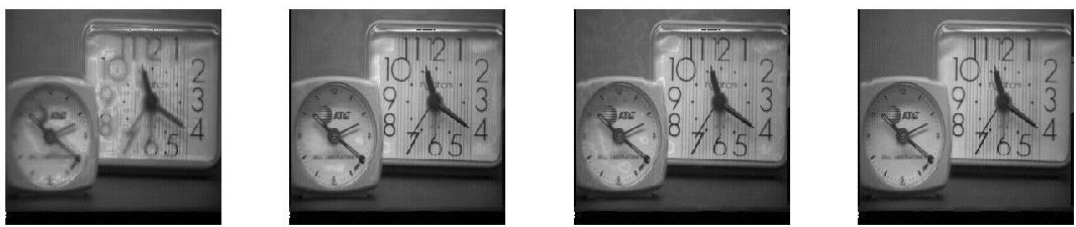

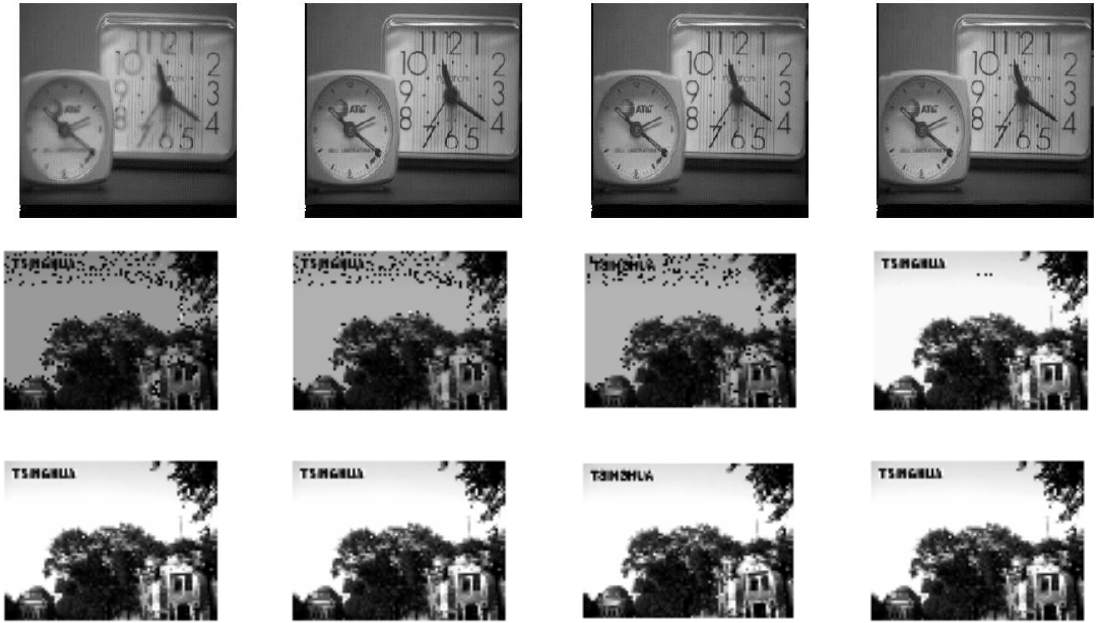

Fig.4 Fusion results using different coefficient (from left to right the selection of measure is variance, spatial frequency, SML and EOL respectively. The first and third lines use DC - PCNN method, the second and fourth rows use the improved DC - PCNN method.)

Table 1 and Table 2 show the comparison of the focus measure by CE, RMSE, PSNR and SSIM obtained on clock images. Table 3 and Table 4 show the comparison of the Tsinghua images. In Table 1 and Table 3, different measures are used in DC-PCNN model, the CE and RMSE calculated by using EOL are smaller than the other three measures, and the PSNR and SSIM are larger, so EOL measure used in DC-PCNN model is the best of the four measures. In Table 2 and Table 4, different measures are used in improved DC-PCNN model, the CE and RMSE which are calculated by using SML are smaller than the other three measures, and the PSNR and SSIM are larger, so the SML measure is the best, but the results by using these measures are similar. Fig.4 shows visual effect of the different methods. The fusion results by using Var, SF, SML in DC-PCNN model are bad, but using EOL in DC-PCNN model is good. While good visual effect is obtained in improved DC-PCNN model by using all the four measures, and the method using SML in improved DC-PCNN model is the best.

**4.2 Experiments of the proposed fusion algorithm**

Groups of source images and reference images are given in Fig.5.

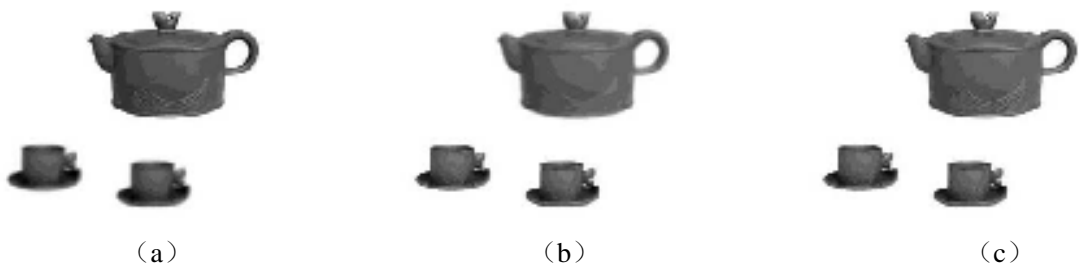

（a） （b） （c）

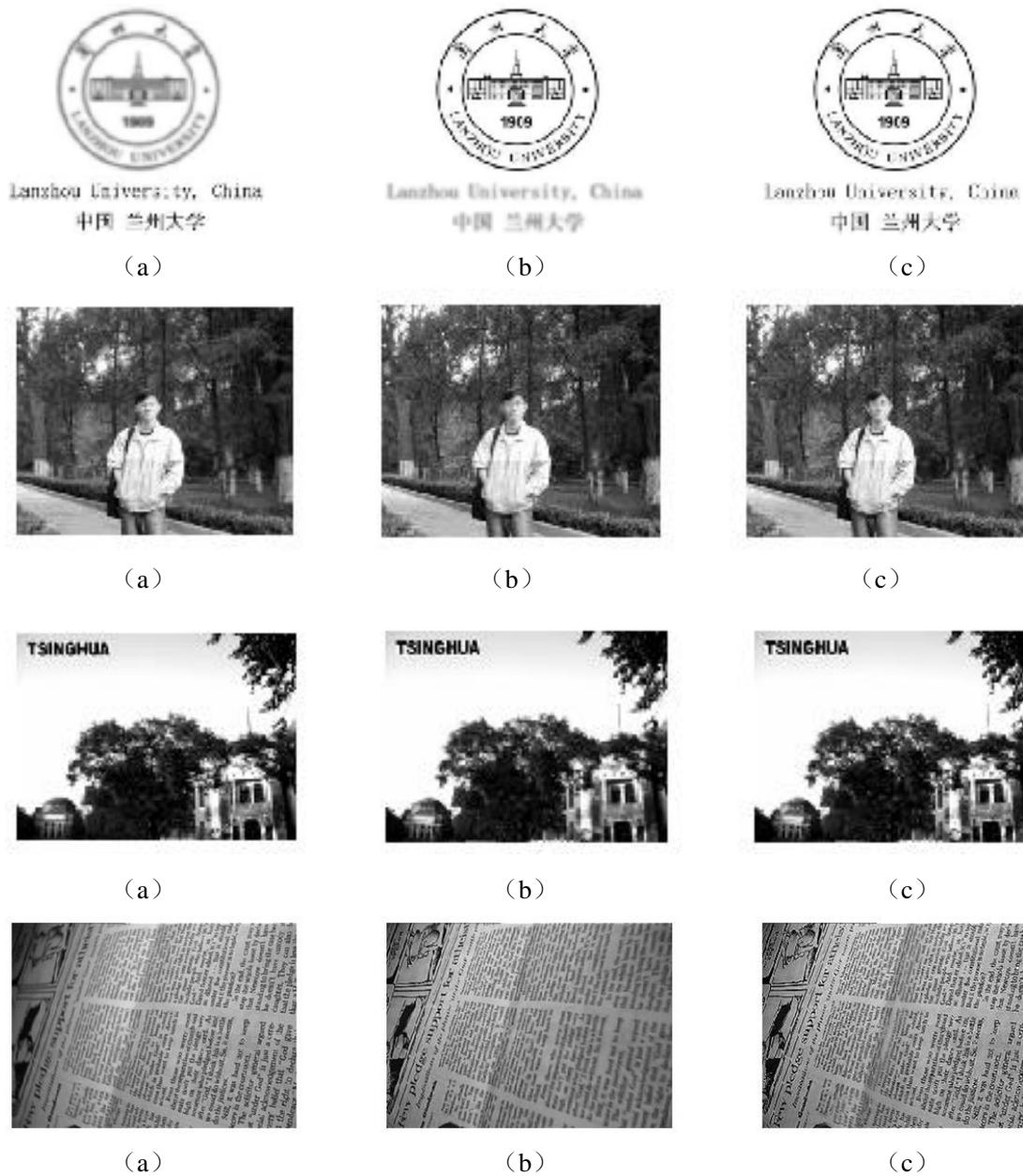

(a)　　　　　　　　　　(b)　　　　　　　　　　(c)

Fig.5　（a）and（b）source images，（c）reference image

The fusion images are given in Fig.6.

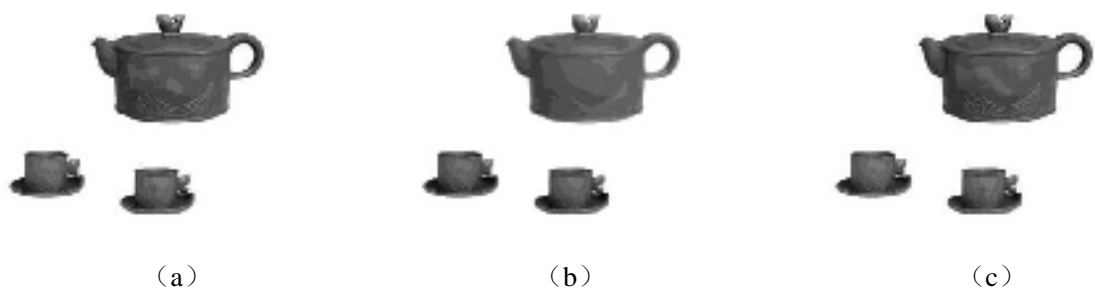

(a)　　　　　　　　　　(b)　　　　　　　　　　(c)

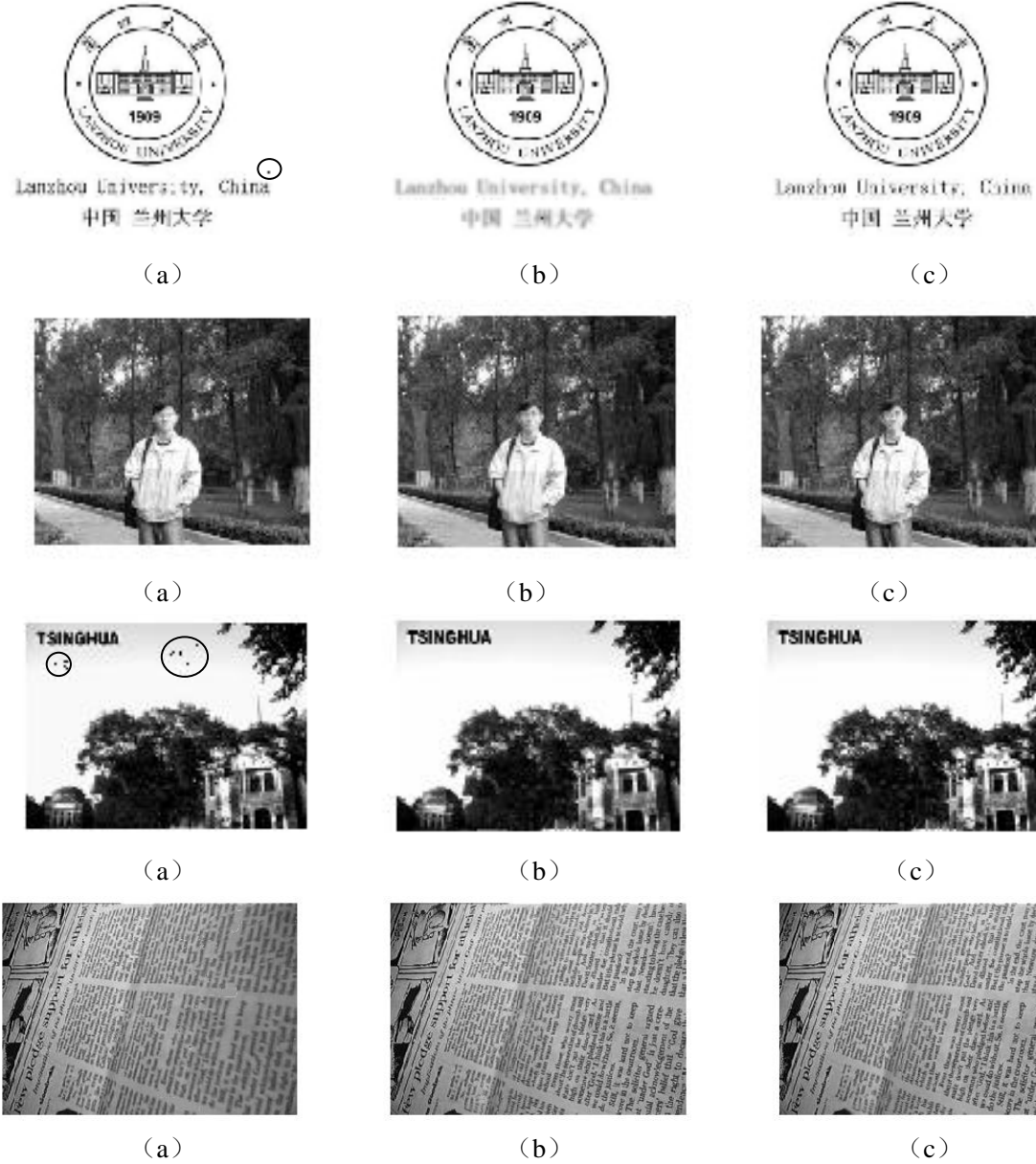

（a） （b） （c）

（a） （b） （c）

（a） （b） （c）

（a） （b） （c）

Fig. 6　The fused images　(a) Wang's method　(b) Huang's method　(c) Our method

Table 5. Evaluation of different fusion methods for the first group of images.

|  | CE | RMSE | PSNR | SSIM | time(s) |
| --- | --- | --- | --- | --- | --- |
| Huang's method | 0.0851 | 5.4660 | 33.3774 | 0.9571 | 51.685061 |
| Wang's method | 0.0060 | 7.9087 | 30.1687 | 0.9479 | 3.620189 |
| Our method | 0.0072 | 6.6889 | 31.6238 | 0.9621 | 3.372264 |

Table 6. Evaluation of different fusion methods for the second group of images.

|  | CE | RMSE | PSNR | SSIM | time(s) |
| --- | --- | --- | --- | --- | --- |
| Huang's method | 0.0336 | 15.8227 | 24.1452 | 0.9243 | 52.068122 |
| Wang's method | 0.0505 | 4.1263 | 35.8197 | 0.9856 | 3.596921 |
| Our method | 0.0505 | 2.8018 | 39.1822 | 0.9931 | 3.423152 |

Table 7. Evaluation of different fusion methods for the third group of images.

|  | CE | RMSE | PSNR | SSIM | time(s) |
|---|---|---|---|---|---|
| Huang's method | 0.0045 | 7.4320 | 30.7087 | 0.8984 | 57.323707 |
| Wang's method | 0.0049 | 7.2511 | 30.9227 | 0.9005 | 4.228810 |
| Our method | 0.0049 | 7.2193 | 30.9609 | 0.9012 | 4.150508 |

Table 8. Evaluation of different fusion methods for the fourth group of images.

|  | CE | RMSE | PSNR | SSIM | time(s) |
|---|---|---|---|---|---|
| Huang's method | 0.0048 | 5.7900 | 32.8772 | 0.9883 | 25.683482 |
| Wang's method | 1.2824 | 8.6508 | 29.3897 | 0.9587 | 1.516650 |
| Our method | 0.0031 | 0.5446 | 53.4088 | 0.9994 | 1.470787 |

Table 9. Evaluation of different fusion methods for the fifth group of images.

|  | CE | RMSE | PSNR | SSIM | time(s) |
|---|---|---|---|---|---|
| Huang's method | 0.1328 | 42.3428 | 15.5952 | 0.3751 | 71.448478 |
| Wang's method | 0.0271 | 27.5612 | 19.3248 | 0.7335 | 4.928975 |
| Our method | 0.0263 | 27.5406 | 19.3313 | 0.7338 | 4.896799 |

Table 5 through Table 9 show that the SSIM values obtained with our method is larger than the other methods. Few abnormal data like CE value in Table 6 through Table 7 and RMSE, PSNR in Table 5, show the Huang's method [11] is better than our method, but this does not affect the merits of our methods, the other criteria of the Huang's method [11] is worse than our method, especially the computational time criterion, the Huang's method [11] is the worst one. Huang's method [11] must employ the PCNN model twice that lead to the lower algorithmic efficiency than others. Wang's method [13] performs one DC - PCNN model that greatly improves the algorithmic efficiency, this can be seen the time column in Table 5 through Table 9. Wang's method [13] improves the image fusion quality, but under some conditions, there are some image fusion spots, such as the second group and the fourth group of images in Fig.6, which requires further processing to the fused image. However, for our method, the objective evaluation criteria are good in Table 5 through Table 9 and the visual effect is well in Fig.6. Furthermore, the computational complexity is less than Wang's method [13].

Although the advantages of the proposed algorithm are exhibited apparently for the task of multi-focus image fusion, there are defects of the proposed model unfortunately for other tasks of image fusion. In order not to mislead readers, six sets of images are adopted to demonstrate the applications of the improved method. The first three groups are multi-focus images, which are suitable for our method. The next three groups are infrared images and visible images, CT image and MR image. However, the fused results using our method for these cases are not so good. These results show the defects of our method.

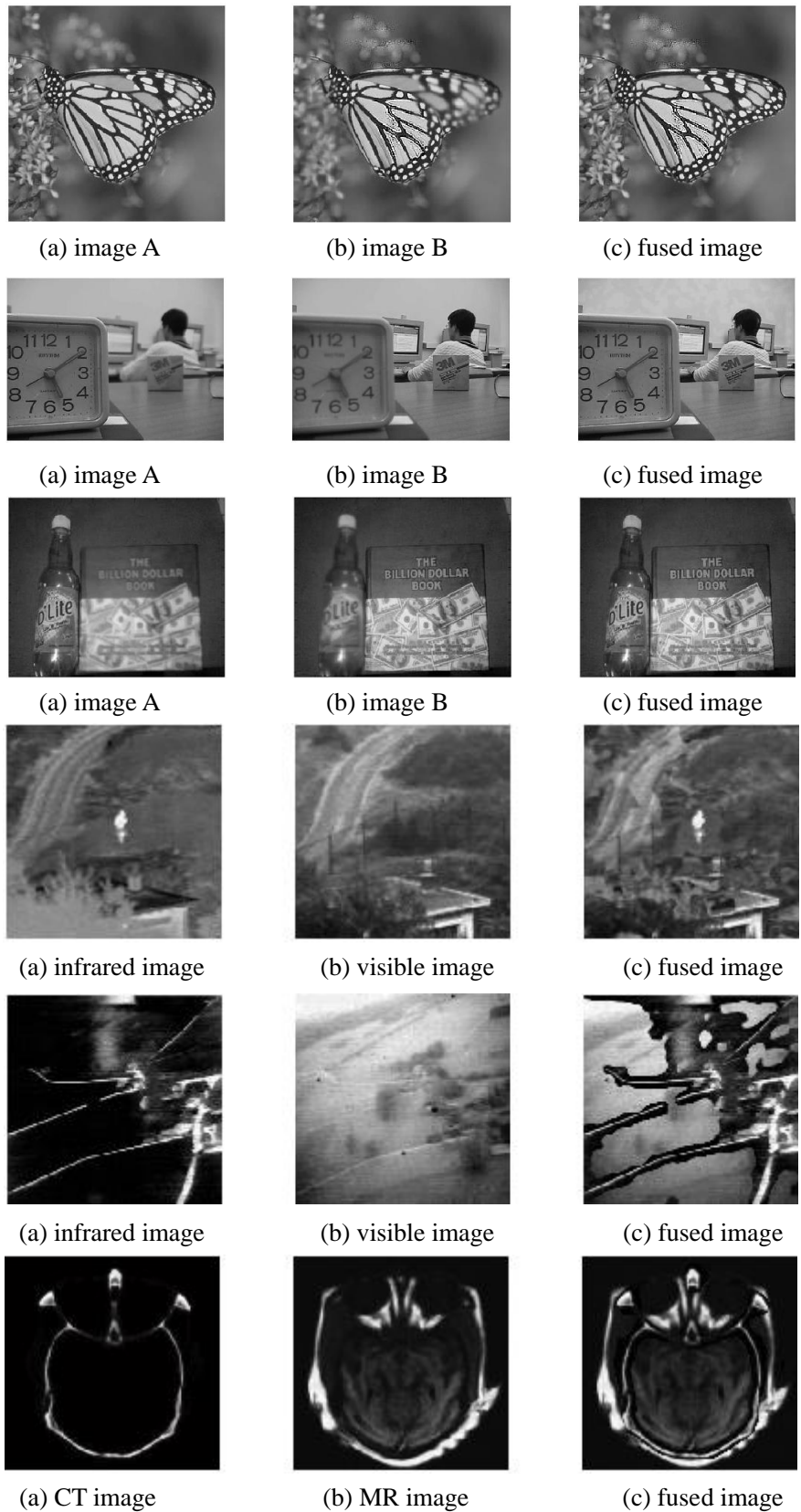

Fig. 7. The applications of the proposed method for different cases.

There are two reasons behind the phenomena. The first reason may be from the fact that there are many parameters in PCNN-like model. However, the values of these parameters in this paper is not optimal in any sense, which leads to unsatisfactory results in image fusion task other than multi-focus image fusion. Fortunately, thanks to the good work of literature [13] and [11], appropriate values of parameters of PCNN-like model have been found for the task of multi-focus image fusion. Our method benefits a lot from [13] and [11] in choosing values of parameters. In this context, we believe our method is very suitable for multi-focus image fusion. The second reason is that the choice of appropriate weighting coefficient of PCNN-like model is a challenging task. The weighting coefficient of the proposed model is obtained for multi-focus image in this paper. There is no way to find an appropriate weighing coefficient of the PCNN-like model to suit for all types of image at present. As a recent future work, we will use global optimization algorithm, say particle swarm optimization (PSO) algorithm [18], to find optimal parameters of PCNN-like models, which may make a small step toward the universality of the proposed method.

## 5. Conclusions

An IDC-PCNN model is proposed, and an improved fusion algorithm based on the improved model is developed. The process of image fusion in the improved method is simpler than that of previous methods. Usually, more PCNN models are employed and more multiplication operates are calculated in the previous methods. However, addition operator instead of multiplication operator is used in the proposed method to implement the multi-focus image. The performance shows that our method is better than the existing methods, whether visual effect or objective evaluation criteria. Furthermore, the computational efficiency is improved. Obviously, lots of works are needed for further research, such as the further improvement of the model, and the selection of the optimal parameters of the model, etc. The optimization model and the choice of appropriate weighting coefficient to be applicable to other types of image are our main work in future.

**Acknowledgements**

This work was supported in part by the following projects: 111 Project of Chinese Ministry of Education (Grant No. B12018), Key Grant Project of Chinese Ministry of Education (Grant No.: 311024), National Natural Science Foundation of P. R. China (Grant No.: 60973094, 61103128), Natural Science Foundation of Jiang Xi Province (No. 20114BAB201022).